\documentclass[10pt,twocolumn,letterpaper]{article}

\usepackage[pagenumbers]{cvpr}

\newcommand{\Lruler}{\mathcal{L}_{\mathrm{ruler}}}
\newcommand{\Ldyn}{\mathcal{L}_{\mathrm{dyn}}}

\usepackage{multirow}
\usepackage{colortbl} 
\usepackage{arydshln}
\definecolor{tabbest}{HTML}{A5A5A5}
\definecolor{tabsecond}{HTML}{D8D8D8}
\newcommand{\gbest}{\cellcolor{tabbest}}
\newcommand{\gsecond}{\cellcolor{tabsecond}}
\definecolor{overallbest}{HTML}{5B9BD5}
\definecolor{overallsecond}{HTML}{BDD7EE}
\newcommand{\obest}{\cellcolor{overallbest}}
\newcommand{\osecond}{\cellcolor{overallsecond}}
\definecolor{backbonebest}{HTML}{8ed873}
\definecolor{backbonesecond}{HTML}{D9F2D0}
\newcommand{\bbest}{\cellcolor{backbonebest}}
\newcommand{\bsecond}{\cellcolor{backbonesecond}}
\definecolor{heldbest}{HTML}{F4CBAB}
\definecolor{heldsecond}{HTML}{F9E5D5}
\newcommand{\hbest}{\cellcolor{heldbest}}
\newcommand{\hsecond}{\cellcolor{heldsecond}}
\DeclareRobustCommand{\swatch}[1]{\textcolor{#1}{\rule[-0.12em]{0.9em}{0.8em}}}
\definecolor{profileteal}{HTML}{168A88}
\definecolor{profileblue}{HTML}{456D94}
\definecolor{profileink}{HTML}{23344A}
\newcommand{\profilefinding}[2]{%
  \par\smallskip\noindent
  \begingroup\setlength{\fboxsep}{5pt}%
  \colorbox{profileteal!7}{%
    \parbox{\dimexpr\linewidth-2\fboxsep\relax}{%
      \small\color{profileink}\textbf{\textcolor{profileteal}{Finding #1.}}~#2}}%
  \endgroup\par\nobreak\smallskip}
\newcommand{\profiledesign}[1]{%
  \par\smallskip\noindent
  \begingroup\setlength{\fboxsep}{5pt}%
  \colorbox{profileblue!7}{%
    \parbox{\dimexpr\linewidth-2\fboxsep\relax}{%
      \small\color{profileink}\textbf{\textcolor{profileblue}{Design implication.}}~#1}}%
  \endgroup\par\smallskip}

\definecolor{cvprblue}{rgb}{0.21,0.49,0.74}
\usepackage[pagebackref,breaklinks,colorlinks,allcolors=cvprblue]{hyperref}

\def\confName{CVPR}
\def\confYear{2027}

\title{DOCK: \underline{D}ual-Stream-Inspired Self-Supervised \underline{O}bject-\underline{C}entric Representations Grounded in Robot \underline{K}inematics}

\author{Pengju Liu\textsuperscript{1} \qquad Caiwen Ding\textsuperscript{1}\\
\textsuperscript{1}University of Minnesota\\
{\ttfamily\small liu03486@umn.edu}\\}

\begin{document}
\maketitle
\begin{abstract}
Latent-space planners select actions by comparing the embedding of a predicted future image with that of a goal image.
This assumes that distance in the embedding reflects how far the scene is from the goal.
We profile this assumption with simulator-executed futures, which removes prediction error.
For six frozen DINO and V-JEPA encoders, feature distance barely tracks the object's distance to its goal: the rank correlation is at most 0.11, against 0.87 for object-mask centroids.
It is dominated by background and robot patches, and frozen-feature readouts capture the arm far better than the object.
Inspired by the ventral (\emph{what}) and dorsal (\emph{where/how}) visual streams, we propose DOCK, which learns an object representation whose changes agree with measured physical motion.
A grasped object moves with the gripper, so forward kinematics on the joint angles gives a proxy for the object's displacement without annotation.
From 9{,}951 held-object transport trajectories in 21 simulated kitchen scenes, we mine 404K displacement pairs and train three components: an object token pooled from patch features (\emph{what}), a \emph{spatial ruler} that makes differences of readout coordinates equal to measured displacements (\emph{where}), and a dynamics predictor over object tokens (\emph{how}).
Used as a planning cost, the learned representation places the object in 15 of 21 tasks, as many as an oracle that reads the true object position, while mask geometry reaches 10 and frozen feature distance 9.
It still solves the most tasks when the unseen robot arm is visible, and its ranking transfers to held-out layouts with unseen object instances and categories without any adaptation.
\end{abstract}

\section{Introduction}
\label{sec:intro}

A world model lets an agent evaluate an action before taking it: the agent predicts the outcome and compares it with the outcome it wants~\cite{ha2018worldmodels,lecun2022path}.
Self-supervised encoders such as DINO~\cite{caron2021dino,oquab2024dinov2,simeoni2025dinov3} and V-JEPA~\cite{bardes2024vjepa,assran2025vjepa2} have made this practical without labels, and recent planners act directly in their feature space.
They encode a goal image, predict the features of candidate futures, and execute the action sequence whose outcome lies closest to the goal~\cite{zhou2025dinowm,assran2025vjepa2,terver2025jepawms}.

\begin{figure}[t]
    \centering
    \includegraphics[width=\linewidth]{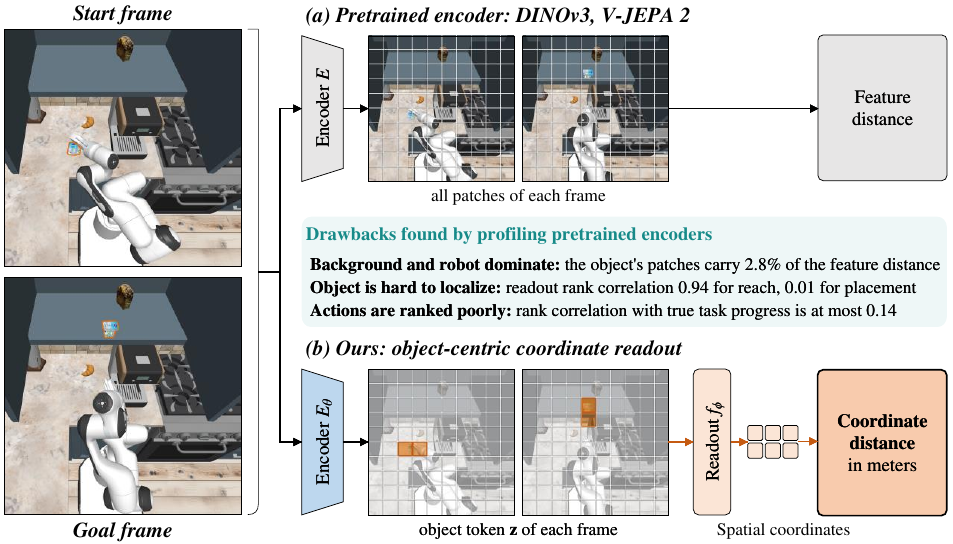}
    
    \caption{\textbf{Feature distance is not a ruler for object location.}
    \textbf{(a)} Planners with pretrained encoders compare all patch features of the two frames, a distance dominated by background and robot.
    \textbf{(b)} We read the object's coordinates from an object token and compare them in meters as the planning cost.}
    \label{fig:teaser}
\end{figure}

This pipeline assumes that the distance between two image embeddings measures progress toward the goal.
For manipulation, current world models have two drawbacks that put this assumption in doubt.
\textbf{(Drawback 1) Task-agnostic relevance:} JEPA-style methods~\cite{assran2023ijepa}, such as V-JEPA~\cite{bardes2024vjepa,assran2025vjepa2}, learn by predicting latent visual representations, aiming to retain meaningful information while suppressing irrelevant detail.
However, what is relevant depends on the downstream task: object placement requires precise spatial localization of the object, yet the object covers only a few image patches and is easily ignored next to the much larger background and robot arm.
\textbf{(Drawback 2) Action-free pretraining:} Encoder pretraining is action-free and does not explicitly model how actions affect future states, which leaves open whether the learned content and appearance features provide sufficient spatial precision.
Exposing and removing these drawbacks raises two challenges.
\textbf{(Challenge 1) Diagnosing the representation in isolation:} A planner can fail because its predictor is wrong or because its cost does not measure the task, and closed-loop success rates do not tell the two apart; the arm also moves with the object it holds, so a cost may follow the arm rather than the object.
\textbf{(Challenge 2) Metric supervision without annotation:} Placement needs a representation that singles out one small object and measures how far and in which direction an action moves it, but current representations largely ignore the object itself, and object-pose or depth labels are costly to collect.

To address these challenges, we present a series of profiling experiments that evaluate the representation alone, and a self-supervised training framework that teaches the representation where an object is and how actions move it, using only the robot's own measured motion.
Our contributions are twofold, as summarized below.

\begin{itemize}[leftmargin=*,itemsep=2pt,topsep=3pt, label={$\triangleright$}]

\item \textbf{A profiling evaluation of current representations (\cref{sec:profiling}).}
In a placement benchmark with exactly simulated futures, the goal-image feature distance of six frozen DINO and V-JEPA encoders is almost blind to where the manipulated object is (\cref{fig:teaser}): its rank correlation with the true object-to-goal distance is at most 0.11, against 0.87 for mask centroids.
It is dominated by background and robot patches, and frozen-feature readouts capture the arm far better than the object.

\item \textbf{DOCK, a self-supervised dataset and training framework (\cref{sec:method}).}
The framework is inspired by the dual-stream of the primate visual system~\cite{mishkin1983pathways,goodale1992pathways}: the ventral pathway for content (\emph{what}) and the dorsal pathway for location and action (\emph{where/how}).
An \emph{object-centric representation} of pretrained features supplies the \emph{what}; a \emph{spatial ruler} that reads out the object's metric location adds the \emph{where}, and \emph{dynamics prediction} of the object's motion under the robot's actions adds the \emph{how} (\cref{fig:framework}).
The ruler and the predictor are trained without annotation on 9{,}951 held-object transport trajectories, using only the gripper motion that forward kinematics computes from joint angles.
As a planning cost, the learned representation places the object in 15 of 21 tasks, as many as an oracle that reads the true object position, against at most 10 for mask geometry and frozen feature distances.

\end{itemize}

\section{Related Work}
\label{sec:related}

\textbf{Visual representations for planning.}
Latent planners predict future features of self-supervised encoders such as DINO~\cite{caron2021dino,oquab2024dinov2,simeoni2025dinov3} and V-JEPA~\cite{bardes2024vjepa,assran2025vjepa2,murlabadia2026vjepa2_1} and minimize their distance to a goal image~\cite{zhou2025dinowm,terver2025jepawms}; others learn visual rewards from value pretraining, temporal distance or demonstrations~\cite{ma2023vip,ma2023liv,hartikainen2020ddl,srinivas2018upn}.
We test whether feature distance tracks physical progress and learn a metric cost instead.
\textbf{Spatial and object-centric representations.}
Probes and 3D-aware fine-tuning study the geometry in pretrained features~\cite{elbanani2024probing3d,man2024lexicon3d,yue2024fit3d}, and object-centric methods rely on slots, object dynamics, descriptors, keypoints or masks~\cite{locatello2020slot,seitzer2023dinosaur,kipf2020cswm,wu2023slotformer,ferraro2023focus,florence2018don,manuelli2019kpam,shi2024pocr}.
We train a mask-pooled object token for metric location readout and action-conditioned prediction.
\textbf{Self-supervision from motion and interaction.}
Ego-motion~\cite{agrawal2015moving,jayaraman2015egomotion} and robot interaction~\cite{agrawal2016poke,pinto2016supersizing,jang2018grasp2vec} supervise visual learning, and measured camera velocity~\cite{guizilini2020packnet} resolves the scale ambiguity of self-supervised depth~\cite{zhou2017sfmlearner,godard2019monodepth2}.
We use gripper displacement from forward kinematics as a metric proxy for a held object's displacement.

\section{Profiling Current Representations}
\label{sec:profiling}
\begin{figure*}[t]
    \centering
    \includegraphics[width=\textwidth]{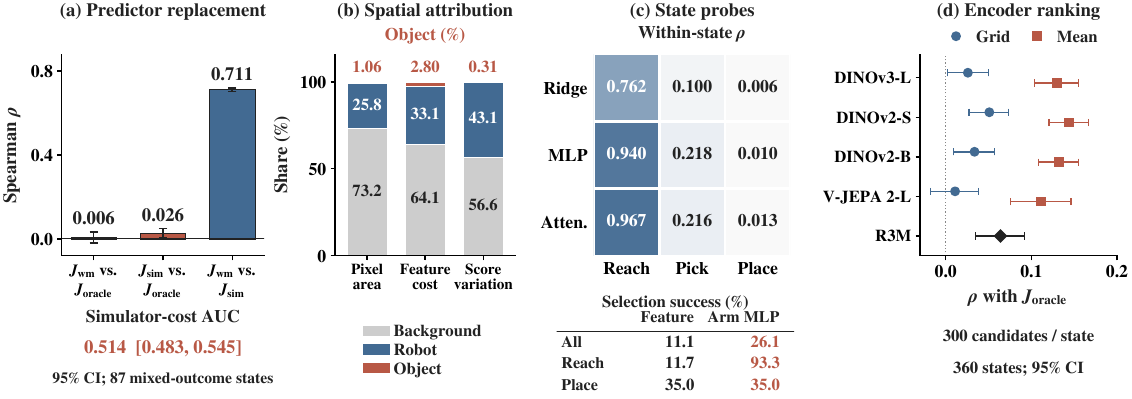}
    \caption{\textbf{Profiling frozen representations} on 360 states with 300 simulator-executed candidates each; (a--c) use DINOv3-L.
    \textbf{(a)} Feature-cost and oracle rank correlations.
    \textbf{(b)} Regional shares of pixel area, feature cost, and score variation.
    \textbf{(c)} Readout correlations and candidate-selection success; place compares separately decoded candidate/goal object positions.
    \textbf{(d)} Feature--oracle correlations across encoders. Whiskers: 95\% bootstrap intervals.}
    \label{fig:profiling}
\end{figure*}
We test whether feature distance ranks physically better actions across six manipulation task families.
For each of 360 states, we execute 300 candidate action sequences from the same saved simulator state and render their outcomes, removing learned-predictor rollout error from scoring.
The full-grid cost is
\begin{equation}
    c_{\mathrm{feat}}(I_c,I_g)=\frac{1}{ND}\sum_{u=1}^{N}
    \bigl\|E(I_c)_u-E(I_g)_u\bigr\|_2^2,
    \label{eq:feature_cost}
\end{equation}
where $N$ and $D$ are the patch count and feature dimension.
We report mean within-state Spearman correlation $\rho$: 1 indicates identical rankings and 0 indicates no monotonic agreement.
Protocols appear in \cref{sec:supp_profiling}.

\subsection{Replacing the predictor with the simulator}
\label{sec:profile_predictor}

\textbf{Experiment.}
We compare feature costs on the world model's predicted features ($J_{\rm wm}$) and on images rendered after executing the same actions in the simulator ($J_{\rm sim}$).
As a physical reference, the \emph{oracle cost} $J_{\rm oracle}$ sums task-specific normalized errors from true simulator states: hand--object distance for reach, object--goal distance for place, and hand--object distance plus the additional height needed to meet the 5\,cm lift target for pick.
Composite tasks sum the applicable terms (\cref{sec:supp_profile_predictor}).
The two feature-cost rankings correlate at $\rho(J_{\rm wm},J_{\rm sim})=0.711$, but their correlations with $J_{\rm oracle}$ are only $0.006$ and $0.026$, respectively (\cref{fig:profiling}a).
Thus, agreement with the predicted feature cost does not imply agreement with physical task progress.
Binary-success AUC measures how often a successful candidate receives a lower cost than an unsuccessful one: 0.5 is chance and 1 is perfect discrimination.
For $J_{\rm sim}$, the mean AUC is $0.514$ over the 87 states containing both outcomes (95\% interval: $[0.483,0.545]$), indicating near-chance discrimination of successful candidates.

\profilefinding{1}{Even with true simulator rollouts, goal-image feature distance does not reliably identify better actions.}

\profiledesign{We evaluate the learned spatial cost on simulator-executed candidate rollouts to isolate scoring quality from predictor error (\cref{sec:setup}).}
\subsection{Spatial attribution of the cost}
\label{sec:profile_attribution}

\textbf{Experiment.}
We split each DINOv3 patch's squared feature difference among background, robot, and object according to the fraction of its pixels in each region, measured by simulator segmentation masks.
\emph{Pixel area} is the fraction of image pixels occupied by a region; \emph{feature cost} is its fraction of the summed patch cost.
\emph{Score variation} measures how the region's cost changes with the total cost across candidates, using the signed covariance share defined in \cref{sec:supp_profile_attribution}.
Background and robot positions account for 64.12\% and 33.09\% of cost magnitude, leaving 2.80\% at object positions.
The object's covariance share is only 0.31\%, so the object hardly changes the score (\cref{fig:profiling}b).

\profilefinding{2}{Most of the full-image cost accumulates at background and robot token positions, motivating an object-focused representation.}

\profiledesign{We pool features over the manipulated object's mask to form the token used by both the spatial readout and dynamics predictor (\cref{sec:object_token}).}

\subsection{Arm and object readouts}
\label{sec:profile_readout}

\textbf{Experiment.}
We freeze DINOv3 and fit ridge and MLP readouts on whole-image mean-pooled features, and an attentive readout on the full token grid.
Reach and pick probes predict hand--object distance and object lift from candidate images.
For placement, we apply the same position readout separately to the candidate and goal images, then compute the Euclidean distance between their predicted 3D object positions.
The MLP attains within-state correlations of $0.940$, $0.218$, and $0.010$ on pure reach, pick, and place; ridge and attention yield $0.006$ and $0.013$ on place (\cref{fig:profiling}c; protocols in \cref{sec:supp_profile_original_readouts}).
The large reach--place gap suggests an arm-based proxy: the readout may estimate approach from arm configuration without accurately localizing the object that has to be placed.

\profilefinding{3}{Strong reach decoding but weak placement decoding suggests that the readout captures arm configuration more readily than object position.}

\profiledesign{We train the visual representation on arm-free images while deriving displacement supervision from robot kinematics (\cref{sec:dataset}).}

\subsection{Candidate correlation across encoders}
\label{sec:profile_ranking}

\textbf{Experiment.}
On the same candidate banks, we compare DINOv2-S/B, DINOv3-L, V-JEPA~2-L, and R3M~\cite{nair2022r3m} by the rank correlation between feature distance and the task-specific oracle cost, using grid, pooled, and available CLS features.
No tested variant exceeds $\rho=0.144$; \cref{fig:profiling}d shows the L2 comparisons.
We also evaluate six frozen encoders on the separate 21-task placement benchmark used in \cref{sec:experiments}, with 300 simulator-generated candidates per task.
In arm-free images, full-image and object-pooled costs reach at most $0.111$ and $0.346$, versus $0.868$ for mask-centroid distance.
On matched visible candidates, DINOv3 object pooling remains far below the mask-centroid distance ($0.420$ vs.\ $0.870$ in rank correlation).

\profilefinding{4}{Across the tested encoders, appearance distance provides weak candidate ordering, motivating a displacement-grounded spatial cost.}

\profiledesign{We train the encoder and readout with metric displacement supervision, then use readout distance as the planning cost (\cref{sec:ruler,sec:planning}).}
\section{DOCK}
\label{sec:method}

Inspired by shared early processing and the specialized ventral and dorsal pathways in human vision~\cite{mishkin1983pathways,goodale1992pathways}, we learn an object representation that plays the role of the ventral pathway (\emph{what}), with a spatial ruler (\emph{where}) and an action-conditioned dynamics predictor (\emph{how}) that play the role of the dorsal pathway.
Held-object transport supplies the motion supervision, and the learned ruler distance provides the metric planning cost at test time (\cref{fig:framework}).


\subsection{Dataset Collection}
\label{sec:dataset}
\begin{figure*}[t]
    \centering
    \includegraphics[width=\linewidth]{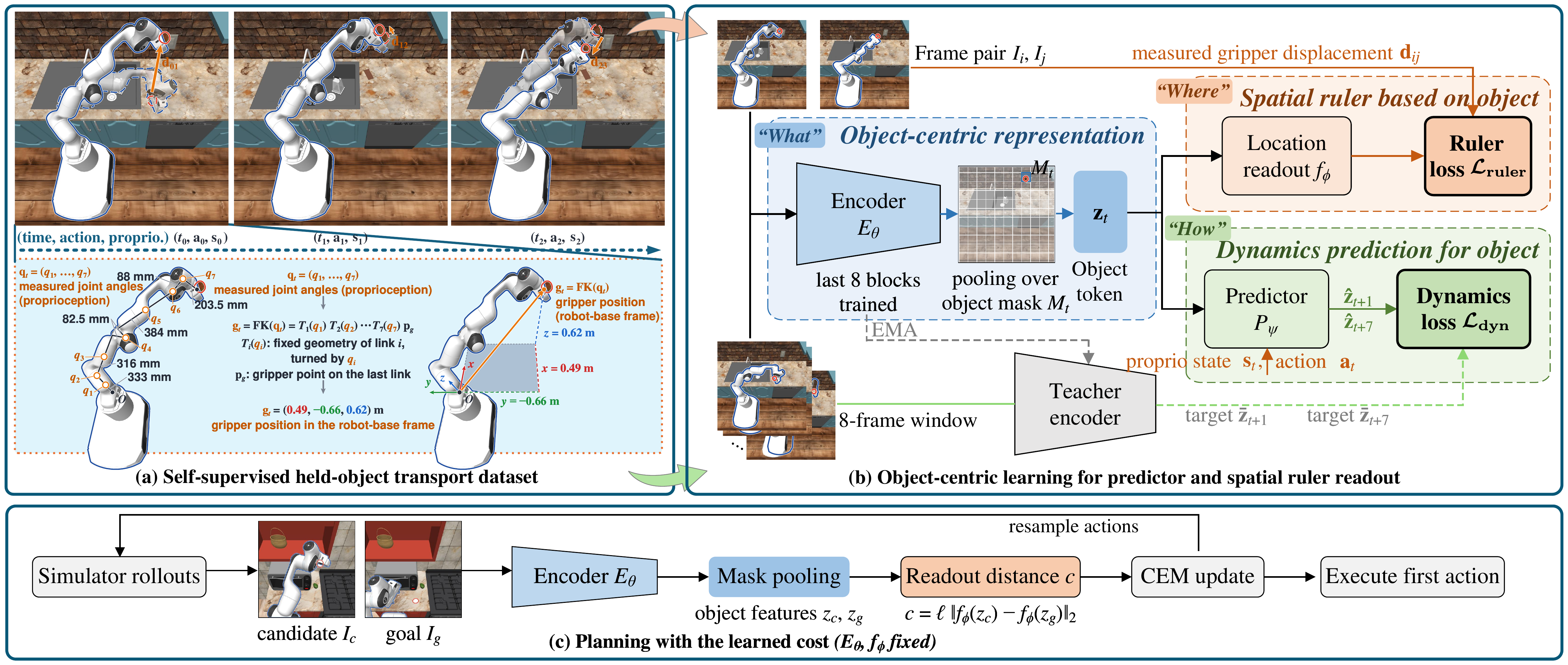}
    \caption{\textbf{Object-centric representation inspired by dual-stream.}
    \textbf{(a) Data collection.} Held-object transport supplies displacement labels from joint angles and forward kinematics.
    \textbf{(b) Joint training.} Mask-pooled object tokens (\emph{what}) support a displacement-supervised spatial ruler (\emph{where}) and an action-conditioned dynamics predictor (\emph{how}), with EMA-teacher targets from eight-frame windows.
    \textbf{(c) Planning.} With the encoder and readout fixed, ruler distance scores simulator rollouts against the goal.}
    \label{fig:framework}
\end{figure*}

\paragraph{Kinematic supervision.}
A person carrying an object can feel how far the hand has moved.
A robot has the same signal: while it holds an object, its joint angles determine the gripper's motion, which gives us displacement supervision without annotation (\cref{fig:framework}a).
For arm joint angles $\mathbf{q}_t\in\mathbb{R}^7$, forward kinematics gives the gripper position $\mathbf{g}_t=\mathrm{FK}(\mathbf{q}_t)\in\mathbb{R}^3$ in the robot-base frame.
The displacement label between frames $i$ and $j$ is
\begin{equation}
    \mathbf{d}_{ij}=\mathbf{g}_j-\mathbf{g}_i .
    \label{eq:ruler_label}
\end{equation}
A grasped object moves with the gripper, so under a stable grasp with little gripper rotation, $\mathbf{d}_{ij}$ serves as a proxy for the object's displacement; metric scale comes from the robot's link geometry.

\paragraph{Held-object transport.}
We collect 9{,}951 held-object transport trajectories in the 21 RoboCasa~\cite{nasiriany2024robocasa} scenes of the benchmark, yielding 3.09 million valid frames (\cref{fig:framework}a).
Each trajectory starts from a saved post-pickup state and moves the arm while maintaining the grasp, with the robot base and camera fixed.
Trajectories include \emph{random walks, persistent and directed transport, reverse paths, expert paths, rotations and surface sweeps}, with additional runs expanding spatial and motion coverage (\cref{sec:supp_dataset}).
Each step records a $256\times256$ image, an object mask, joint angles and contact information.
The same state is rendered with and without the robot; training uses the arm-free view to \emph{discourage decoding object position through the arm}.

From stable-grasp segments without object--environment contact, we construct two training sets.
\emph{Displacement pairs}, labeled by \cref{eq:ruler_label}, require a visible object in both frames, at least 5\,mm of gripper translation, and at most $2^\circ$ of rotation.
Sampling across eight length bands yields 404{,}397 training pairs spanning 0.5\,cm to beyond 90\,cm.
\emph{Temporal windows} contain eight visible frames at 0.25\,s intervals, yielding 82{,}405 training windows; they permit gripper rotation within a stable grasp.

Whole trajectories are split 80/10/5/5 into training, validation, readout-fitting and test partitions.
Transfer to unseen layouts, object instances and categories is evaluated separately (\cref{sec:generalization}).
\Cref{tab:dataset_pairs} summarizes pair counts by displacement length and partition.
\Cref{sec:supp_dataset} details the collection protocol, pair and window filtering, partitions and coverage statistics of the collected data.

\begin{table}[t]
\centering
\caption{\textbf{Displacement pairs by length.} All partitions pass the same pair filter. Training is capped per band and scene; the other partitions keep every eligible pair. \emph{Fit} and \emph{Fit-val.}\ are the two parts of the readout-fitting partition.}
\label{tab:dataset_pairs}
\footnotesize
\setlength{\tabcolsep}{3.5pt}
\renewcommand{\arraystretch}{1.04}
\begin{tabular}{@{}lrrrrr@{}}
\toprule
Length (cm) & Train & Val. & Fit & Fit-val. & Test \\
\midrule
$[0.5,2)$ & 63,000 & 633 & 188,681 & 24,584 & 197,537 \\
$[2,5)$ & 63,000 & 633 & 149,365 & 19,966 & 160,875 \\
$[5,10)$ & 63,000 & 633 & 107,307 & 14,839 & 114,698 \\
$[10,20)$ & 63,000 & 628 & 73,208 & 10,244 & 77,478 \\
$[20,35)$ & 63,000 & 627 & 37,562 & 5,556 & 39,891 \\
$[35,60)$ & 56,450 & 588 & 16,703 & 3,403 & 20,055 \\
$[60,90)$ & 31,293 & 354 & 2,917 & 474 & 5,556 \\
$\geq 90$ & 1,654 & 0 & 300 & 0 & 594 \\
\midrule
Total & \textbf{404,397} & \textbf{4,096} & \textbf{576,043} & \textbf{79,066} & \textbf{616,684} \\
\bottomrule
\end{tabular}
\end{table}

\subsection{Training Framework}
\label{sec:training}
\label{sec:object_token}
\label{sec:ruler}
\label{sec:dynamics}
\label{sec:planning}

\paragraph{Object-centric representation (What).}
A pretrained ViT-L/16 encoder $E_\theta$, either DINOv3~\cite{simeoni2025dinov3} or V-JEPA~2~\cite{assran2025vjepa2}, maps the full image $I_t$ to a $16\times16$ grid of patch features.
For V-JEPA~2, the image is repeated twice to form one tubelet.
With $w_{t,u}\in[0,1]$ denoting the fraction of patch $u$ covered by the simulator object mask $M_t$, the object token is
\begin{equation}
    \mathbf{z}_t =
    \frac{\sum_u w_{t,u}\, E_\theta(I_t)_u}{\sum_u w_{t,u}}
    \in\mathbb{R}^{1024}.
    \label{eq:object_pool}
\end{equation}
The mask acts only at pooling, and no geometric augmentation changes the object's image position.
The same object token provides a compact visual input to both the location readout and dynamics predictor.
This shared representation is analogous to early visual processing that supports specialized downstream pathways.

\paragraph{Spatial ruler based on object (Where).}
The spatial ruler learns from pairs of frames $(I_i,I_j)$ separated by a measured displacement $\mathbf{d}_{ij}$.
A three-layer perceptron $f_\phi:\mathbb{R}^{1024}\rightarrow\mathbb{R}^3$ maps their object tokens to coordinates whose difference should match this displacement:
\begin{equation}
    \Lruler =
    \mathbb{E}_{(i,j)}\!\left[
    h_\delta\!\left(
    \left\|f_\phi(\mathbf{z}_j)-f_\phi(\mathbf{z}_i)
    -\frac{\mathbf{d}_{ij}}{\ell}\right\|_2\right)\right],
    \label{eq:ruler_loss}
\end{equation}
Here $\ell=0.10$\,m sets the coordinate scale, and $h_\delta$ is the Huber function~\cite{huber1964robust} with $\delta=0.05$, corresponding to 5\,mm.
The loss penalizes displacement error equally in all directions and updates both the readout and encoder.
Only coordinate differences are supervised, so the readout learns a metric frame with a free origin and never needs an absolute object position as a label.

\paragraph{Dynamics prediction for object (How).}
A causal transformer predicts the next object token from visual history, proprioception and achieved actions:
\begin{equation}
    \hat{\mathbf{z}}_{t+1}=P_\psi\bigl(\mathbf{z}_{\le t},\,\mathbf{s}_{\le t},\,\mathbf{a}_{\le t}\bigr).
    \label{eq:predictor}
\end{equation}
The seven arm joint angles $\mathbf{q}_t$ shown in \cref{fig:framework}a determine the gripper pose through forward kinematics.
We concatenate its position (3 values) and orientation quaternion (4) with the two finger-joint positions to form the proprioceptive state $\mathbf{s}_t\in\mathbb{R}^9$.
The achieved action $\mathbf{a}_t\in\mathbb{R}^7$ contains the translation, rotation vector and change in the sum of finger-joint positions between frames.
Targets $\bar{\mathbf{z}}_{t+1}$ come from a stop-gradient exponential-moving-average (EMA) teacher~\cite{grill2020byol,assran2023ijepa}.
For each eight-frame window, teacher forcing (TF) predicts the next token from object tokens encoded from the observed frames up to the current step.
Autoregressive rollout (AR) starts from the first observed token and feeds each prediction back as the next visual input.
Both modes use the same proprioceptive states and achieved actions and are supervised at all seven transitions:
\begin{equation}
    \begin{aligned}
    \mathcal{L}_m &= \frac{1}{7BK}
    \sum_{b=1}^{B}\sum_{t=0}^{6}\sum_{k=1}^{K}
    \left(\frac{\hat{z}^{m}_{b,t+1,k}-\bar{z}_{b,t+1,k}}{\sigma_k}\right)^{\!2},\\
    \Ldyn &= \tfrac{1}{2}
    \left(\mathcal{L}_{\mathrm{TF}}+\mathcal{L}_{\mathrm{AR}}\right),
    \end{aligned}
    \label{eq:dynamics_loss}
\end{equation}
Here $m\in\{\mathrm{TF},\mathrm{AR}\}$, $B$ is the number of windows, $K=1024$ is the token dimension, and $\sigma_k$ is the batch standard deviation of teacher-token dimension $k$.

\paragraph{Joint training.}
We jointly train the readout, predictor, and last eight encoder blocks with the final normalization layer.
The losses are normalized by stop-gradient running means:
\begin{equation}
    \mathcal{L} =
    \frac{\Lruler}{m_{\mathrm{ruler}}+\epsilon}
    +\frac{\Ldyn}{m_{\mathrm{dyn}}+\epsilon},
    \label{eq:joint_loss}
\end{equation}
The running means have a decay of 0.99.
Each update uses 256 pairs and 64 windows with uniformly sampled scenes.
For the main models, we use AdamW~\cite{loshchilov2019adamw} for at most 20{,}000 updates and select checkpoints by validation loss; fixed-endpoint ablations and implementation details are described in \cref{sec:ablation,sec:supp_implementation}, respectively.
\begin{table*}[t]
\centering
\caption{\textbf{Main results on the 21 placement tasks} (protocol in \cref{sec:setup}). \swatch{overallbest}\,/\,\swatch{overallsecond}: best and second-best value in each column, excluding the oracle; \swatch{tabbest}\,/\,\swatch{tabsecond}: best and second-best frozen-feature cost for each encoder.}
\label{tab:main_results}
\small
\setlength{\tabcolsep}{4.5pt}
\renewcommand{\arraystretch}{1.04}
\begin{tabular}{@{}lrrrrrrr>{\columncolor{white}[\tabcolsep][0pt]}r@{}}
\toprule
& \multicolumn{4}{c}{Closed-loop planning} & \multicolumn{4}{c}{Decision-0 ranking} \\
\cmidrule(lr){2-5}\cmidrule(lr){6-9}
& \multicolumn{2}{c}{Arm-free} & \multicolumn{2}{c}{Robot-visible} & \multicolumn{2}{c}{Arm-free} & \multicolumn{2}{c}{Robot-visible} \\
\cmidrule(lr){2-3}\cmidrule(lr){4-5}\cmidrule(lr){6-7}\cmidrule(lr){8-9}
Cost & Dist.\ (cm)$\downarrow$ & Succ.$\uparrow$ & Dist.\ (cm)$\downarrow$ & Succ.$\uparrow$ & $\rho\uparrow$ & Regret (mm)$\downarrow$ & $\rho\uparrow$ & Regret (mm)$\downarrow$ \\ \hline
\midrule
\multicolumn{9}{l}{\textit{Frozen feature distance}} \\ 
DINOv3-L, all patches & 41.23 & \gsecond 8/21 & 48.72 & \gsecond 3/21 & \gsecond 0.045 & \gbest 127.36 & \gbest 0.200 & \gbest 172.64 \\
DINOv3-L, goal mask & \gsecond 34.33 & \gsecond 8/21 & 49.40 & 2/21 & $-$0.002 & 156.72 & 0.064 & 188.35 \\
DINOv3-L, candidate mask & 51.61 & 2/21 & \gsecond 47.71 & \gsecond 3/21 & $-$0.078 & 224.51 & $-$0.012 & 235.34 \\
DINOv3-L, both masks & 44.75 & 3/21 & 48.37 & 1/21 & $-$0.011 & 200.40 & $-$0.032 & \gsecond 180.96 \\
DINOv3-L, pooled object & \gbest 30.01 & \gbest 9/21 & \gbest 46.45 & \gbest 5/21 & \gbest 0.346 & \gsecond 146.03 & \gsecond 0.114 & 225.92 \\ \hdashline
V-JEPA~2-AC-g, all patches & 47.50 & \gsecond 6/21 & \gsecond 41.38 & \gbest 3/21 & 0.009 & 219.93 & \gbest 0.153 & 213.35 \\
V-JEPA~2-AC-g, goal mask & \gsecond 39.63 & 3/21 & \gbest 40.92 & \gbest 3/21 & \gsecond 0.018 & \gsecond 150.10 & 0.083 & \gbest 187.51 \\
V-JEPA~2-AC-g, candidate mask & 54.31 & 2/21 & 49.83 & 1/21 & $-$0.045 & 237.30 & $-$0.035 & 244.94 \\
V-JEPA~2-AC-g, both masks & 42.76 & 4/21 & 47.35 & \gbest 3/21 & $-$0.009 & 176.82 & $-$0.044 & 210.75 \\
V-JEPA~2-AC-g, pooled object & \gbest 33.53 & \gbest 8/21 & 47.29 & \gbest 3/21 & \gbest 0.256 & \gbest 141.86 & \gbest 0.153 & \gsecond 188.83 \\
V-JEPA~2-AC-g, released predictor$^\dagger$ & 55.69 & 1/21 & 54.67 & 0/21 & \textemdash & \textemdash & \textemdash & \textemdash \\ \hline
\midrule
\multicolumn{9}{l}{\textit{Mask geometry}} \\
Mask centroid & 19.43 & 9/21 & \osecond 23.15 & 8/21 & 0.868 & 50.21 & \obest 0.452 & \obest 77.53 \\
Mask centroid + size & 17.86 & 10/21 & 26.89 & 7/21 & 0.837 & \osecond 22.59 & 0.213 & 88.88 \\ \hline
\midrule
\multicolumn{9}{l}{\textbf{\textit{DOCK (ours)}}} \\
\textbf{DINOv3-L} & \obest 15.48{\scriptsize$\,\pm\,$0.05} & \obest 15/21 & \obest 20.72{\scriptsize$\,\pm\,$0.36} & \obest 12/21 & \obest 0.915 & \obest 12.12 & \osecond 0.435 & 86.97 \\
\textbf{V-JEPA~2-L} & \osecond 16.13{\scriptsize$\,\pm\,$0.51} & \obest 15/21 & 24.01{\scriptsize$\,\pm\,$0.26} & \osecond 9/21 & \osecond 0.894 & 26.00 & 0.395 & \osecond 85.82 \\ \hline
\midrule
\multicolumn{9}{l}{\textit{Reference}} \\
True object distance (oracle) & 12.56 & 15/21 & 12.56$^*$ & 15/21$^*$ & 1.000 & 0.00 & 1.000 & 0.00 \\
\bottomrule
\end{tabular}
\par\smallskip
\begin{minipage}{\textwidth}\footnotesize
$\dagger$: candidate futures from the released V-JEPA~2-AC predictor instead of the simulator. $*$: the oracle uses no images, so one set of episodes serves both renderings. $\pm$: standard deviation over three evaluation repeats; success counts are identical across repeats.
\end{minipage}
\end{table*}

\paragraph{Planning cost.}
For candidate and goal images of the same scene and object, the learned cost estimates their object-to-goal distance in meters:
\begin{equation}
    c(I_c,I_g)=\ell\,
    \bigl\|f_\phi(\mathbf{z}_c)-f_\phi(\mathbf{z}_g)\bigr\|_2 ,
    \label{eq:location_cost}
\end{equation}
This replaces the feature distance of \cref{eq:feature_cost} in CEM~\cite{rubinstein1999cem} (\cref{sec:setup}).
Candidate and goal are images of the same scene, so the free origin of the readout cancels in the cost.
Our cost-isolation evaluations generate candidate futures with the simulator.
We evaluate both the jointly trained (\emph{co-trained}) readout and a \emph{fresh} readout fitted on the readout-fitting partition with the encoder frozen, enabling a common readout protocol across training objectives.

\section{Experiments}
\label{sec:experiments}

\subsection{Setup}
\label{sec:setup}
\paragraph{Benchmark and metrics.}
We evaluate on the 21 placement tasks, rendered arm-free (AF) and robot-visible (RV) from identical states; a task succeeds if the object ends within 15\,cm of its goal.
\emph{Decision-0 ranking} scores a fixed bank of 300 simulated candidate futures per task and reports the Spearman correlation $\rho$ between cost and true object-to-goal distance, and the regret (true distance of the lowest-cost candidate minus that of the best).
\emph{Closed-loop planning} runs CEM~\cite{rubinstein1999cem} for six decisions of eight iterations each on simulator-generated futures~\cite{zhu2020robosuite,todorov2012mujoco}, which isolates the cost from prediction error, and reports the mean final distance and the number of solved tasks.
Displacement error and further protocol details are in \cref{sec:supp_benchmark}.
\textbf{Models.}
We adapt the last eight blocks (\emph{Last~8}) of DINOv3 and V-JEPA~2 ViT-L/16 on arm-free images only.
Unless stated otherwise, the planning cost (\cref{eq:location_cost}) uses the \emph{co-trained} readout, trained jointly with the encoder, rather than a \emph{fresh} one fitted afterwards with the encoder fixed.
Each model has one training seed; reported deviations cover three planning repeats and measure execution variability only.
\textbf{Baselines.}
Frozen feature distance (\cref{eq:feature_cost}) is computed over \emph{all patches}; with patch differences weighted by the object's coverage of each patch in the goal image (\emph{goal mask}, where the object should be), in the candidate image (\emph{candidate mask}, where it is) or by their sum (\emph{both masks}); and between mask-pooled object features (\emph{pooled object}, \cref{eq:object_pool}).
We also compare with V-JEPA~2-AC planning with its released predictor, mask geometry (centroid pixel distance and its pinhole extension with mask-area depth), and an oracle that uses the true object-to-goal distance as its cost.

\subsection{Main results}
\label{sec:main_results}

\begin{table*}[t]
\centering
\caption{\textbf{Which loss adapts the encoder?} Encoders adapted for 20{,}000 updates, each with a fresh readout. \swatch{backbonebest}\,/\,\swatch{backbonesecond}: best and second-best value within each backbone. Regret in mm.}
\label{tab:loss_ablation}
\footnotesize
\setlength{\tabcolsep}{3.8pt}
\renewcommand{\arraystretch}{1.04}
\begin{tabular}{@{}llrrrrrrrrr>{\columncolor{white}[\tabcolsep][0pt]}r@{}}
\toprule
& & \multicolumn{2}{c}{Error (mm)$\downarrow$} & \multicolumn{2}{c}{Ranking, AF} & \multicolumn{2}{c}{Ranking, RV} & \multicolumn{2}{c}{Planning, AF} & \multicolumn{2}{c}{Planning, RV} \\
\cmidrule(lr){3-4}\cmidrule(lr){5-6}\cmidrule(lr){7-8}\cmidrule(lr){9-10}\cmidrule(lr){11-12}
Backbone & Adaptation loss & Test & Cube & $\rho\uparrow$ & Regret$\downarrow$ & $\rho\uparrow$ & Regret$\downarrow$ & Dist.\ (cm)$\downarrow$ & Succ.$\uparrow$ & Dist.\ (cm)$\downarrow$ & Succ.$\uparrow$ \\
\midrule
DINOv3-L & None (frozen encoder) & 23.95 & 21.53 & 0.862 & 65.07 & 0.341 & 115.28 & 18.78{\scriptsize$\,\pm\,$0.06} & 10/21 & 29.23{\scriptsize$\,\pm\,$0.05} & 7/21 \\
 & Dynamics only & 25.40 & 24.06 & 0.850 & 66.73 & 0.348 & 128.79 & 23.75{\scriptsize$\,\pm\,$0.18} & 7/21 & 29.90{\scriptsize$\,\pm\,$0.01} & 7/21 \\
 & Ruler only & \bbest 14.82 & \bbest 15.68 & \bbest 0.888 & \bsecond 26.86 & \bbest 0.425 & \bsecond 82.47 & \bsecond 17.05{\scriptsize$\,\pm\,$0.09} & \bsecond 13/21 & \bsecond 22.38{\scriptsize$\,\pm\,$0.06} & \bbest 11/21 \\
 & Ruler + dynamics & \bsecond 15.19 & \bsecond 15.75 & \bsecond 0.885 & \bbest 23.15 & \bsecond 0.419 & \bbest 54.65 & \bbest 15.75{\scriptsize$\,\pm\,$0.09} & \bbest 15/21 & \bbest 21.00{\scriptsize$\,\pm\,$0.18} & \bsecond 10/21 \\
\midrule
V-JEPA~2-L & None (frozen encoder) & 26.40 & 24.47 & 0.847 & 59.22 & 0.363 & 116.85 & \bbest 18.59{\scriptsize$\,\pm\,$0.04} & \bsecond 11/21 & 32.27{\scriptsize$\,\pm\,$0.51} & \bsecond 7/21 \\
 & Dynamics only & 28.12 & 25.16 & 0.804 & 81.72 & 0.350 & 130.78 & 26.86{\scriptsize$\,\pm\,$0.17} & 8/21 & 27.34{\scriptsize$\,\pm\,$0.15} & \bsecond 7/21 \\
 & Ruler only & \bbest 18.51 & \bsecond 18.20 & \bsecond 0.875 & \bsecond 38.02 & \bbest 0.429 & \bbest 75.06 & 19.06{\scriptsize$\,\pm\,$0.03} & \bsecond 11/21 & \bbest 21.81{\scriptsize$\,\pm\,$0.66} & \bbest 9/21 \\
 & Ruler + dynamics & \bsecond 18.85 & \bbest 16.95 & \bbest 0.876 & \bbest 35.92 & \bsecond 0.427 & \bsecond 100.22 & \bsecond 18.90{\scriptsize$\,\pm\,$0.02} & \bbest 13/21 & \bsecond 26.53{\scriptsize$\,\pm\,$0.22} & \bsecond 7/21 \\
\bottomrule
\end{tabular}
\end{table*}

\begin{table*}[t]
\centering
\caption{\textbf{Encoder adaptation and readout.} \emph{Frozen}: only the readout and predictor are trained; \emph{Last~8}: the last eight encoder blocks are trained as well. \swatch{backbonebest}\,/\,\swatch{backbonesecond}: best and second-best value within each backbone. Regret in mm.}
\label{tab:variants}
\footnotesize
\setlength{\tabcolsep}{3.6pt}
\renewcommand{\arraystretch}{1.04}
\begin{tabular}{@{}lllrrrrrrrrr>{\columncolor{white}[\tabcolsep][0pt]}r@{}}
\toprule
& & & \multicolumn{2}{c}{Error (mm)$\downarrow$} & \multicolumn{2}{c}{Planning, AF} & \multicolumn{2}{c}{Planning, RV} & \multicolumn{2}{c}{Ranking, AF} & \multicolumn{2}{c}{Ranking, RV} \\
\cmidrule(lr){4-5}\cmidrule(lr){6-7}\cmidrule(lr){8-9}\cmidrule(lr){10-11}\cmidrule(lr){12-13}
Backbone & Encoder & Readout & Test & Cube & Dist.\ (cm)$\downarrow$ & Succ.$\uparrow$ & Dist.\ (cm)$\downarrow$ & Succ.$\uparrow$ & $\rho\uparrow$ & Regret$\downarrow$ & $\rho\uparrow$ & Regret$\downarrow$ \\
\midrule
DINOv3-L & Frozen & Co-trained & 23.42 & 24.19 & 17.06{\scriptsize$\,\pm\,$0.05} & 14/21 & 29.96{\scriptsize$\,\pm\,$0.17} & 8/21 & 0.883 & 19.71 & 0.314 & 104.05 \\
 & Frozen & Fresh & 23.95 & 21.53 & 18.78{\scriptsize$\,\pm\,$0.06} & 10/21 & 29.23{\scriptsize$\,\pm\,$0.05} & 7/21 & 0.862 & 65.07 & 0.341 & 115.28 \\
 & Last 8 & Co-trained & \bbest 12.39 & \bbest 15.27 & \bbest 15.48{\scriptsize$\,\pm\,$0.05} & \bbest 15/21 & \bbest 20.72{\scriptsize$\,\pm\,$0.36} & \bbest 12/21 & \bbest 0.915 & \bbest 12.12 & \bbest 0.435 & \bsecond 86.97 \\
 & Last 8 & Fresh & \bsecond 15.17 & \bsecond 15.80 & \bsecond 15.75{\scriptsize$\,\pm\,$0.09} & \bbest 15/21 & \bsecond 21.00{\scriptsize$\,\pm\,$0.18} & \bsecond 10/21 & \bsecond 0.885 & \bsecond 17.26 & \bsecond 0.420 & \bbest 50.34 \\
\midrule
V-JEPA~2-L & Frozen & Co-trained & 21.18 & 24.01 & \bsecond 17.19{\scriptsize$\,\pm\,$0.12} & \bsecond 14/21 & 26.79{\scriptsize$\,\pm\,$0.14} & \bbest 9/21 & \bsecond 0.877 & \bbest 25.88 & 0.363 & 108.44 \\
 & Frozen & Fresh & 26.40 & 24.47 & 18.59{\scriptsize$\,\pm\,$0.04} & 11/21 & 32.27{\scriptsize$\,\pm\,$0.51} & 7/21 & 0.847 & 59.22 & 0.363 & 116.85 \\
 & Last 8 & Co-trained & \bbest 14.54 & \bsecond 19.15 & \bbest 16.13{\scriptsize$\,\pm\,$0.51} & \bbest 15/21 & \bbest 24.01{\scriptsize$\,\pm\,$0.26} & \bbest 9/21 & \bbest 0.894 & \bsecond 26.00 & \bsecond 0.395 & \bbest 85.82 \\
 & Last 8 & Fresh & \bsecond 18.79 & \bbest 16.86 & 18.90{\scriptsize$\,\pm\,$0.02} & 13/21 & \bsecond 26.53{\scriptsize$\,\pm\,$0.22} & 7/21 & 0.876 & 35.92 & \bbest 0.427 & \bsecond 101.26 \\
\bottomrule
\end{tabular}
\end{table*}

\textbf{Planning.}
In arm-free images, the learned cost solves 15 of 21 tasks with both backbones, as many as the oracle (\cref{tab:main_results}), at a mean final distance of 15.5 and 16.1\,cm against the oracle's 12.6\,cm.
Five of the six tasks our models fail also fail with the oracle, which points to the planner rather than the cost; the sixth has a hidden goal object, on which our cost abstains.
Mask geometry solves at most 10 tasks, all of which our models also solve; frozen feature distance solves at most 9 and leaves the object 30\,cm or more from the goal on average, and V-JEPA~2-AC with its own predictor solves only one of the 21 tasks.

\textbf{Ranking.}
The learned costs also order the initial candidates best in arm-free images ($\rho=0.915$ and $0.894$, against $0.868$ for the mask centroid).
The DINOv3 model halves the regret of the best baseline (12.1 against 22.6\,mm for mask centroid + size), while the V-JEPA~2 model stays slightly above it (26.0\,mm).
Frozen feature distance stays far behind ($\rho\le0.346$, regret of at least 127\,mm).

\textbf{Mask pooling.}
Among the frozen-feature costs, pooling over the object mask, as our object token does (\cref{eq:object_pool}), is the most effective use of the mask in arm-free images: it gives the highest correlation ($\rho=0.346$ and $0.256$, against at most $0.045$) and the best planning, whereas mask-weighted patch distances do not consistently improve on all patches; the other frozen encoders behave similarly (\cref{tab:eval_ranking_frozen,tab:supp_planning_frozen}).
Pooling alone does not make the feature metric, however: with the same pooling, our trained encoder and readout raise the correlation $\rho$ from $0.346$ to $0.915$ for DINOv3 in arm-free images.

\textbf{Robot-visible images.}
Our models never see the robot in training, yet with the arm rendered the DINOv3 model still solves 12 tasks at 20.7\,cm and the V-JEPA~2 model 9, against 8 at 23.2\,cm for the mask centroid.
Decision-0 ranking is weak for every cost here, and the mask centroid even ranks the initial candidates slightly better than our models ($\rho=0.452$ against $0.435$ and $0.395$); yet over six decisions the DINOv3 model solves four more tasks.
Frozen all-patch distance instead correlates better when the arm is visible ($\rho$ from $0.045$ to $0.200$ for DINOv3), consistent with a cost dominated by robot and background patches (\cref{sec:profile_attribution}).

\subsection{Ablations}
\label{sec:ablation}

\begin{table*}[t]
\centering
\caption{\textbf{Transfer to held-out instances and categories}, in unseen kitchen layouts. Succ.: tasks solved in each of the three repeats; $\rho$: decision-0 ranking correlation. \swatch{heldbest}\,/\,\swatch{heldsecond}: best and second-best value in each column; \swatch{tabbest}\,/\,\swatch{tabsecond}: best and second-best frozen-feature cost. Final distance and regret are in \cref{tab:supp_heldout_instances,tab:supp_heldout_categories}.}
\label{tab:heldout}
\small
\setlength{\tabcolsep}{5.5pt}
\renewcommand{\arraystretch}{1.04}
\begin{tabular}{@{}lccccccc>{\columncolor{white}[\tabcolsep][0pt]}c@{}}
\toprule
& \multicolumn{4}{c}{Instance-held-out (10 tasks)} & \multicolumn{4}{c}{Category-held-out (9 tasks)} \\
\cmidrule(lr){2-5}\cmidrule(lr){6-9}
& \multicolumn{2}{c}{Arm-free} & \multicolumn{2}{c}{Robot-visible} & \multicolumn{2}{c}{Arm-free} & \multicolumn{2}{c}{Robot-visible} \\
\cmidrule(lr){2-3}\cmidrule(lr){4-5}\cmidrule(lr){6-7}\cmidrule(lr){8-9}
Cost & Succ.$\uparrow$ & Ranking $\rho\uparrow$ & Succ.$\uparrow$ & Ranking $\rho\uparrow$ & Succ.$\uparrow$ & Ranking $\rho\uparrow$ & Succ.$\uparrow$ & Ranking $\rho\uparrow$ \\
\midrule
\multicolumn{9}{l}{\textit{Frozen feature distance}} \\
DINOv3-L, all patches & \gsecond 2,\,2,\,2 & 0.046 & \gbest 1,\,1,\,1 & \gbest 0.184 & \gsecond 2,\,2,\,2 & 0.052 & 0,\,0,\,0 & 0.049 \\
DINOv3-L, pooled object & \gbest 4,\,4,\,4 & \gbest 0.291 & \gbest 1,\,1,\,1 & 0.014 & \gbest 3,\,3,\,3 & \gbest 0.482 & \gbest 2,\,2,\,2 & \gbest 0.228 \\
V-JEPA~2-L, all patches & 1,\,2,\,2 & \gsecond 0.192 & 0,\,0,\,0 & \gsecond 0.104 & 1,\,1,\,1 & 0.052 & 0,\,0,\,0 & 0.032 \\
V-JEPA~2-L, pooled object & 1,\,1,\,1 & 0.050 & 0,\,0,\,0 & 0.001 & 1,\,1,\,1 & \gsecond 0.164 & \gsecond 1,\,1,\,1 & \gsecond 0.080 \\
\midrule
\multicolumn{9}{l}{\textit{DOCK (ours), trained on the original 21 scenes only}} \\
DINOv3-L, co-trained readout & \hsecond 5,\,4,\,4 & \hsecond 0.796 & \hsecond 2,\,2,\,2 & \hbest 0.488 & \hsecond 5,\,5,\,5 & \hbest 0.740 & 2,\,2,\,2 & \hbest 0.402 \\
DINOv3-L, fresh readout & \hsecond 4,\,4,\,5 & \hbest 0.843 & 1,\,1,\,1 & \hsecond 0.440 & \hbest 6,\,6,\,6 & \hsecond 0.729 & \hbest 3,\,3,\,3 & \hsecond 0.393 \\
V-JEPA~2-L, co-trained readout & 3,\,3,\,2 & 0.697 & \hbest 3,\,3,\,3 & 0.320 & 4,\,4,\,4 & 0.635 & 1,\,1,\,1 & 0.354 \\
V-JEPA~2-L, fresh readout & \hbest 6,\,5,\,6 & 0.783 & \hsecond 2,\,2,\,2 & 0.410 & 3,\,3,\,3 & 0.656 & \hbest 3,\,3,\,3 & 0.353 \\
\bottomrule
\end{tabular}
\par\smallskip
\begin{minipage}{\textwidth}\footnotesize
One arm-free and two robot-visible goals of the instance-held-out tasks are hidden; those tasks abstain and count as failures, and $\rho$ is averaged over the other 9 and 8 tasks.
\end{minipage}
\end{table*}

\textbf{Ruler versus dynamics.}
\Cref{tab:loss_ablation} adapts the encoder with each loss alone and with both (ours), for 20{,}000 updates and with the same fresh-readout protocol.
\emph{Test} and \emph{Cube} are the median displacement errors on held-out test pairs and on pairs with the object inside a 10\,cm cube of each scene, which are withheld from training.
Against the frozen encoder, our model lowers the test error from 24.0 to 15.2\,mm for DINOv3 and from 26.4 to 18.9\,mm for V-JEPA~2, within 0.4\,mm of ruler-only, with similar gains inside the cube.
It gives the lowest arm-free regret for both backbones (23.2 and 35.9\,mm) and solves the most arm-free tasks: 15 and 13, against 13 and 11 for ruler-only and at most 11 for the other two.
For DINOv3 it also cuts the robot-visible regret of ruler-only from 82.5 to 54.7\,mm and reaches the lowest final distance in both renderings, while for V-JEPA~2 the robot-visible results favor ruler-only.

\textbf{Encoder adaptation.}
\Cref{tab:variants} compares our adapted encoder with a frozen one, for which only the readout and predictor are trained, and shows that training the encoder as well is necessary.
With the co-trained readout, adaptation nearly halves the displacement error for DINOv3 (23.4 to 12.4\,mm) and cuts it by a third for V-JEPA~2 (21.2 to 14.5\,mm).
It also improves the final distance in every planning setting and the ranking correlation in both renderings, for both backbones.
The gain is largest when the unseen arm is visible, where adaptation raises the DINOv3 count from 8 to 12 tasks and lowers its mean final distance from 30.0 to 20.7\,cm in closed-loop planning.

\textbf{Co-trained versus fresh readout.}
\Cref{tab:variants} also compares the \emph{co-trained} readout with a \emph{fresh} one fitted later with the encoder fixed, which tests whether the metric structure lies in the token or only in its jointly trained head.
The co-trained readout is the better cost: in every configuration it has lower test error and arm-free regret and solves at least as many tasks.
The gap depends on the encoder: on an adapted encoder a fresh readout solves 15 (DINOv3) and 13 (V-JEPA~2) arm-free tasks against 15 with the co-trained one, while on a frozen encoder it loses four and three tasks.
This suggests that adaptation places the metric structure in the token itself, not in a particular readout.

\subsection{Generalization to held-out instances and categories benchmark}
\label{sec:generalization}

We test transfer without any adaptation, readout fitting or calibration on two benchmarks whose target assets and kitchen layouts do not occur in the original 21 scenes: ten \emph{instance-held-out} tasks, five of whose target categories still occur there, and nine \emph{category-held-out} tasks whose target categories are new as well.

\textbf{Unseen instances and layouts.}
The learned costs keep ranking candidates well in new kitchens (\cref{tab:heldout}): in AF, $\rho=0.697$--$0.843$ against at most $0.291$ for frozen feature distance, at about half the regret (48--55 against 91--178\,mm; \cref{tab:supp_heldout_instances}).
Planning gains are smaller and depend on the model: the best learned cost (V-JEPA~2, fresh readout) solves 5--6 of ten tasks against four for the best frozen baseline (pooled DINOv3), while the co-trained V-JEPA~2 readout solves only 2--3.

\textbf{Unseen categories and layouts.}
The advantage holds for unseen categories: $\rho=0.635$--$0.740$ in AF against at most $0.482$, and DINOv3 with a fresh readout solves six of nine tasks, twice as many as pooled frozen DINOv3.
Neither readout is consistently better on held-out tasks, and RV remains hard in both benchmarks: the learned costs solve 1--3 tasks against 0--2 for frozen features, so transfer does not remove the sensitivity to the visible robot.


\section{Conclusion}
\label{sec:conclusion}

Our profiling shows that feature distance, the cost of latent-space planners, fails for object placement: for six frozen encoders it is dominated by background and robot patches and barely tracks the object's distance to its goal.
Inspired by the ventral and dorsal visual streams, DOCK instead learns an object token (\emph{what}), a spatial ruler (\emph{where}) and a dynamics predictor (\emph{how}) from held-object transport, supervised only by forward kinematics.
With simulated futures, its metric cost solves 15 of 21 placement tasks, matching an oracle with the true object position, and still solves the most tasks with the unseen arm visible; its ranking transfers to unseen layouts, instances and categories.



\clearpage
\section*{LLM Usage Statement}
We used large language models (Claude and Codex) to polish the writing of this paper, to write parts of the code, and to launch and schedule the experiments.
The research ideas are our own, and all figures were drawn by the authors.

\section*{Impact Statement}
This paper presents work whose goal is to advance the field of machine learning.
There are many potential societal consequences of our work, none of which we feel must be specifically highlighted here.

{
    \small
    \bibliographystyle{ieeenat_fullname}
    \bibliography{main}
}

\clearpage
\setcounter{page}{1}
\renewcommand*{\thepage}{\texorpdfstring{\arabic{page}}{supp.\arabic{page}}}
\addtocontents{brf}{\protect
\let\supp@savedbackcite\backcite
\def\backcite#1#2{\supp@markcitation#2\supp@savedbackcite{#1}{#2}}
\def\supp@markcitation#1#2#3{\global\@namedef{supp@cited@#3}{}}
\global\let\supp@paperbackref\backrefxxx
\gdef\backrefxxx#1#2#3{%
  \ifcsname supp@cited@#3\endcsname
    \hyperlink{page.supp.#1}{#1}%
  \else
    \supp@paperbackref{#1}{#2}{#3}%
  \fi}
}
\maketitlesupplementary
\appendix

\noindent We will release our code with the camera-ready version of the paper.

\section{Profiling protocols}
\label{sec:supp_profiling}

The comparisons in \cref{fig:profiling} use candidates from the first optimizer iteration at decision 8. For each of the six task families (reach, pick, place, reach--pick, pick--place, and reach--pick--place), six planning seeds contribute ten episodes each. We assign average ranks to tied candidates for Spearman's $\rho$, count ties as half a win for binary-success AUC, and compute confidence intervals with a task-stratified bootstrap.
For the cross-encoder comparison in \cref{fig:profiling}d, R3M supplies one global vector. For V-JEPA~2 we repeat one static frame 64 times and average the patch grid over time.

\subsection{Task-specific oracle costs}
\label{sec:supp_profile_predictor}

Let $d_{ho}$ and $d_{og}$ denote hand--object and object--goal distance, and let $\ell$ denote object lift, all in meters. The normalized error terms are
\begin{equation}
    r=\frac{d_{ho}}{0.20},\qquad
    p=\frac{\max(0,0.05-\ell)}{0.05},\qquad
    q=\frac{d_{og}}{0.15}.
\end{equation}
The oracle scores each candidate as
\begin{equation}
    J_{\rm oracle}=\begin{cases}
        r & \text{reach},\\
        r+p & \text{pick or reach--pick},\\
        q & \text{place},\\
        r+p+q & \text{pick--place or reach--pick--place}.
    \end{cases}
\end{equation}

\setcounter{subsection}{2}
\subsection{Arm and object readout protocols}
\label{sec:supp_profile_original_readouts}

Each pure task contributes 60 states. In each of three folds, we fit the readout on four planning seeds and test on the other two, keeping all candidates from an episode together. We average within-state correlations across the equal-sized folds.
All readouts use unmasked features. The attentive readout uses one learned query over 256 patch tokens of dimension 1,024.

For placement in \cref{fig:profiling}c, we fit a readout to simulator-measured 3D object positions in candidate images. We apply it to each candidate and its goal, then rank the 300 candidates by the distance between their predicted positions and the predicted goal position. The robot is out of contact with the object in 59 of 60 states. The median within-state range of true object--goal distances is 6.7\,mm, so the correlation tests fine differences in distance to the goal. \Cref{tab:variants} reports displacement error for the separate mask-pooled readouts.
For candidate selection in the same panel, a separately fitted MLP selects the sequence with the smallest predicted hand--object distance. Success is the recorded outcome of that sequence in the fixed candidate bank.

\setcounter{subsection}{5}
\subsection{Signed covariance share}
\label{sec:supp_profile_attribution}

For candidate $k$ and patch $u$, let $e_{ku}$ be the squared feature difference summed over feature dimensions. Let $w_{kur}$ be the fraction of the patch covered by region $r$ in the terminal image. The cost assigned to region $r$ and the total cost are
\begin{equation}
    S_{kr}=\sum_u w_{kur}e_{ku},\qquad
    T_k=\sum_u e_{ku}=\sum_r S_{kr}.
\end{equation}
Within each state, the signed share of score variation for region $r$ is
\begin{equation}
    v_r=\frac{\operatorname{Cov}_k(S_{kr},T_k)}{\operatorname{Var}_k(T_k)}.
\end{equation}
\Cref{fig:profiling}b averages these shares over the 360 decision-8 states. The shares sum to one because the regional costs sum to $T_k$. A negative share means the regional cost covaries inversely with the total. These shares refer to token positions. Self-attention allows object pixels to affect tokens over the background.

\section{Dataset details}
\label{sec:supp_dataset}

\paragraph{Trajectory collection.}
Each motion command lasts five control steps at 20\,Hz. Random walks sample translations and small rotations independently. Persistent motion holds a sampled direction for 2--12 decisions while keeping the object in view and within reach. Directed transport varies distance and speed along straight paths, with pauses, returns to the start, or retracing in reverse. Expert paths follow recorded gripper poses with slower, perturbed, and branching variants. For rotation sequences we rotate the gripper, translate at the new orientation, and rotate back. For surface sweeps we lower the object, lift it slightly, and translate horizontally.
We use candidate and goal positions from the original 21-task benchmark to choose spatial coverage targets. We measure coverage on a 5\,cm grid and direct additional trajectories toward sparsely observed cells, starting from nearby saved states with stable grasps when available.
Further runs sample underrepresented displacement lengths and directions, and rotate the gripper to less frequent orientations before translating.
A trajectory ends on a failure such as grasp loss or prolonged contact, or after at most 900 control steps.

\paragraph{Segments and pairs.}
We first remove warm-up frames, frames with contact between the object and the environment, and unreliable trajectory tails.
A segment is a run of frames in which the pose of the object relative to the gripper stays within 15\,mm and $5^\circ$ of its value at the segment start.
Within each segment, pairs in every partition must meet the visibility and motion criteria in \cref{sec:dataset} and show agreement between simulator object displacement and the kinematic target in \cref{eq:ruler_label}.
Training keeps at most 3{,}000 pairs per length band and scene.
Validation is a fixed, approximately scene-balanced subset of 4{,}096 pairs that fills the available bands in round-robin order; it contains no pair of 90\,cm or more.
Readout fitting and testing keep all eligible pairs of their partitions.
We evaluate the placement benchmark on its own candidate--goal pairs without transport-pair screening and measure distances between simulator object positions.

\paragraph{Windows.}
Each window stays within one stable-grasp segment. We cap training windows at 4{,}000 per scene and retain 511 validation windows in total.

\paragraph{Partitions.}
We implement the split in \cref{sec:dataset} by hashing whole trajectories. The \emph{Cube} test measures displacement in a designated 10\,cm cube in each scene (\cref{tab:loss_ablation,tab:variants}). We exclude a pair or window from training if the object is inside the cube at either endpoint. A window can still cross the cube between its endpoints, and 2{,}200 cube-cohort frames occur in dynamics-training windows.
The readout-fitting partition has 655{,}109 pairs.
A fixed rule on scene and trajectory identifiers divides them into 576{,}043 pairs that update a fresh readout and 79{,}066 pairs that are used only for early stopping and for selecting the best readout; all pairs of a trajectory fall into the same part.
This internal validation set is distinct from the 4{,}096 validation pairs that select the encoder checkpoint.

\paragraph{Minibatch sampling.}
Within each sampled scene, we draw uniformly from its retained pairs or windows, with replacement. Pairs and windows are drawn independently, and the sampling probability of a length band is proportional to its retained pair count.
Fresh-readout fitting instead shuffles all fitting pairs in every epoch.

\paragraph{Coverage statistics.}
For each scene, we take the earliest eligible training frame in the first trajectory that contributes a training pair as the reference orientation. We measure the smallest 3D rotation from each retained object's orientation to the reference, counting each frame once per partition. The median training angle ranges from $2.7^\circ$ to $18.5^\circ$ across the 21 scenes.

To measure directional coverage, we normalize each displacement and compare it with 20{,}000 approximately uniformly spaced rays. Across scenes, 97.4\%--100\% of rays lie within $5^\circ$ of an observed training displacement. Across training windows, the median start-to-end displacement is 4.4\,cm (95th percentile 21.2\,cm). Summing the distances traveled over all 35 control steps gives a median path length of 4.9\,cm (95th percentile 26.8\,cm).

\section{Implementation details}
\label{sec:supp_implementation}

\paragraph{Forward kinematics.}
For \cref{eq:ruler_label} we include the Panda wrist and gripper mounts in the kinematic chain from the base link to the gripper site.
Across all 3.43 million recorded frames, the computed and simulator-recorded gripper positions in the base frame agreed to within $10^{-9}$\,m. On a physical arm, the accuracy of this label depends on kinematic calibration.

\paragraph{Preprocessing.}
RGB values are scaled from $[0,255]$ to $[0,1]$ and normalized per channel with the ImageNet mean $(0.485,0.456,0.406)$ and standard deviation $(0.229,0.224,0.225)$.

\paragraph{Location readout.}
The readout $f_\phi$ is a perceptron with layer widths $1024\rightarrow512\rightarrow512\rightarrow3$ and GELU activations.
In \cref{eq:ruler_loss} we use $h_\delta(r)=r^2/2$ for $r<\delta$ and $\delta(r-\delta/2)$ otherwise.

\paragraph{Predictor.}
Starting from the first recorded state in each training window, we integrate the achieved motions to reconstruct the subsequent states. The fingers remain closed during transport. The gripper-action channel in \cref{sec:dynamics} records the change in the sum of two oppositely signed finger-joint positions, so their contributions can cancel as the gripper opens.
Three linear projections map $\mathbf{a}_t$, $\mathbf{s}_t$ and $\mathbf{z}_t$ to 384-dimensional tokens.
A learned time embedding, shared by the three tokens of a frame, and a learned type embedding for motion, state or visual token are added to each.
A transformer with 12 layers, 12 heads and width 384 processes the tokens under a causal mask that admits the current and earlier time steps.
After a final layer normalization, the visual token of each step passes through a linear layer $384\rightarrow1024$ to give $\hat{\mathbf{z}}_{t+1}$.
The teacher encoder is an exponential moving average of the student with decay 0.999.

\paragraph{Dynamics loss.}
We compute the standard deviation $\sigma_k$ in \cref{eq:dynamics_loss} from the $7B$ teacher tokens in each forward pass. We floor the per-dimension variance at $10^{-6}$ before taking its square root and treat the result as a constant when computing gradients.

\paragraph{Loss normalization.}
We initialize each running mean in \cref{eq:joint_loss} with the first minibatch loss and update it after each optimizer step using the mean unnormalized loss for that step. We set $\epsilon=10^{-12}$.

\paragraph{Optimization.}
We use AdamW with learning rates $3\times10^{-5}$ for the encoder and $3\times10^{-4}$ for the readout and the predictor, 500 warm-up updates and cosine decay to 10\% of the peak rate.
Every 500 updates we evaluate the ruler loss on the 4{,}096 validation pairs and log the two dynamics losses on the 511 validation windows.
Only the ruler validation loss selects the checkpoint and controls early stopping, which occurs after six evaluations without improvement once 5{,}000 updates have elapsed.
The frozen-encoder variant uses the same stopping rule. In the fixed-endpoint ablations (\cref{sec:ablation}), planning repeats share one fresh readout per condition. In the held-out evaluation, each repeat uses its corresponding readout from the original data. Both renderings use readouts trained on arm-free images.

\section{Benchmark and evaluation details}
\label{sec:supp_benchmark}

\paragraph{Tasks.}
The 21 tasks comprise 18 pick-and-place tasks between kitchen fixtures and three composite tasks. Each task has one post-pickup start state and one goal from an expert demonstration.

\paragraph{Candidate banks and planner.}
For each task, decision-0 ranking uses a common bank of 300 three-action sequences (0.75\,s each) from the saved start state.
In closed loop, each CEM iteration samples 300 sequences and retains 10 elites. For the cost comparison, the simulator executes candidate actions and each visual cost scores the rendered outcome against the goal. The released-predictor baseline in \cref{tab:main_results} generates its own futures. After the last iteration the robot executes the first action of the elite mean for five control steps.

\paragraph{Held-out protocols.}
The reference inventory contains all targets, containers and distractors in the original 21 scenes: 53 unique assets, 40 categories and 15 layouts.
We use this inventory to define what was unseen during DOCK training (\cref{sec:generalization}). The held-out comparison evaluates DOCK and frozen-feature costs on the same tasks and candidate banks, fixed before evaluation. Category-held-out tasks span five familiar PLACE task classes.

For instance-held-out tasks we check bilateral fingerpad contact, absence of external contact and grasp stability in the preceding frame. Category-held-out tasks also require a 0.5\,s holding test and a saved goal free of external occlusion in both renderings. During evaluation, every cost uses simulator masks to check visibility. An empty candidate mask incurs a large penalty. An empty goal mask triggers a zero-action hold without CEM. This occurs for one arm-free and two robot-visible goals among the ten instance tasks and for no category-held-out goals.

Both splits include fish and eggplant and kitchen layouts 15, 17 and 44. The eggplant tasks use different start and goal frames from one \texttt{PickPlaceCounterToStove} episode. We report each split separately. Ranking correlations average over 9 arm-free and 8 robot-visible instance tasks and all 9 category tasks. Regret and planning include every task in each split.
For each model, the three planning repeats use the same trained encoder, task starts and goals, and CEM seed (1). The standard deviations in \cref{tab:supp_heldout_instances,tab:supp_heldout_categories} are computed across these repeats.

\paragraph{Masks and abstention on the original benchmark.}
The learned ruler and mask-geometry costs penalize candidates with empty masks. One of the 21 goals has an empty mask. On that task these costs select the first candidate for ranking and execute zero actions in closed loop.
The frozen-feature ranking uses the full-image distance when a required mask is empty (\cref{tab:eval_ranking_frozen}). For the matched-visible comparison in \cref{sec:profile_ranking}, each scorer ranks the same visible candidates within a rendering, excluding the task with the empty goal.

\paragraph{Metrics.}
We compute Spearman's $\rho$ within each task and average over tasks with a defined correlation. On the original benchmark this gives 21 tasks for frozen features and 20 for the ruler and mask centroid. Regret and success include all 21.
Object-to-goal distance is measured between body origins.
For displacement error, we convert the difference of the two readout coordinates to meters using $\ell=0.10$\,m and take its Euclidean distance from the measured displacement. We report the median in millimeters, separately for the 616{,}684 test pairs and the 36{,}295 cube pairs (\cref{sec:supp_dataset,tab:loss_ablation,tab:variants}).
Only four initial banks contain a candidate within 15\,cm of the goal. This bounds decision-0 selection success in \cref{tab:eval_ranking_frozen} at 4/21. Closed-loop success is measured after six planning decisions.

\clearpage
\setcounter{page}{7}
\setcounter{table}{14}
\begin{table*}[!tp]
\centering
\caption{\textbf{Frozen-feature ranking on the full 21-task benchmark.} The lower panel adds decision-0 success for the two encoders whose correlations and regrets appear in \cref{tab:main_results}. Bold and underline identify the best and second-best objectives within each encoder, including ties.}
\label{tab:eval_ranking_frozen}
\small
\setlength{\tabcolsep}{5pt}
\renewcommand{\arraystretch}{1.04}
\begin{tabular}{@{}llrrrrrr@{}}
\toprule
& & \multicolumn{3}{c}{Arm-free (AF)} & \multicolumn{3}{c}{Robot-visible (RV)} \\
\cmidrule(lr){3-5}\cmidrule(lr){6-8}
Frozen encoder & Objective & $\rho\uparrow$ & Regret (mm)$\downarrow$ & Success$\uparrow$ & $\rho\uparrow$ & Regret (mm)$\downarrow$ & Success$\uparrow$ \\
\midrule
DINOv2-S & Unmasked & 0.097 & 179.87 & \textbf{2/21} & \textbf{0.231} & \textbf{180.11} & \underline{0/21} \\
 & Goal mask & 0.024 & \textbf{152.72} & \textbf{2/21} & -0.071 & 203.31 & \textbf{1/21} \\
 & Candidate mask & 0.064 & 267.64 & \underline{1/21} & 0.003 & 212.07 & \underline{0/21} \\
 & Both masks & \underline{0.105} & 229.23 & \underline{1/21} & -0.089 & \underline{192.54} & \underline{0/21} \\
 & Pooled object & \textbf{0.252} & \underline{160.86} & \underline{1/21} & \underline{0.088} & 236.55 & \underline{0/21} \\
\midrule
DINOv2-B & Unmasked & 0.111 & 158.56 & \textbf{3/21} & \textbf{0.206} & \textbf{177.80} & \underline{1/21} \\
 & Goal mask & 0.075 & \textbf{119.44} & \underline{2/21} & -0.040 & 229.11 & \textbf{2/21} \\
 & Candidate mask & 0.082 & 208.79 & 1/21 & 0.012 & 220.84 & \underline{1/21} \\
 & Both masks & \underline{0.118} & 193.46 & \underline{2/21} & -0.123 & \underline{199.56} & \textbf{2/21} \\
 & Pooled object & \textbf{0.301} & \underline{146.88} & 1/21 & \underline{0.075} & 200.43 & \underline{1/21} \\
\midrule
V-JEPA~2-L $\times2$ & Unmasked & 0.037 & 180.93 & \textbf{3/21} & \textbf{0.143} & \textbf{201.86} & \textbf{0/21} \\
 & Goal mask & \underline{0.040} & \textbf{150.82} & \underline{2/21} & -0.059 & \underline{220.23} & \textbf{0/21} \\
 & Candidate mask & -0.019 & 237.06 & 1/21 & -0.034 & 222.99 & \textbf{0/21} \\
 & Both masks & 0.003 & 176.49 & \underline{2/21} & -0.063 & 267.86 & \textbf{0/21} \\
 & Pooled object & \textbf{0.213} & \underline{175.22} & \underline{2/21} & \underline{0.038} & 242.80 & \textbf{0/21} \\
\midrule
V-JEPA~2.1-L (image) & Unmasked & 0.064 & \underline{143.94} & \textbf{3/21} & \textbf{0.213} & 198.48 & \textbf{1/21} \\
 & Goal mask & \underline{0.076} & 169.86 & \underline{2/21} & 0.090 & \textbf{157.58} & \textbf{1/21} \\
 & Candidate mask & 0.038 & 186.47 & 1/21 & 0.019 & 202.98 & \textbf{1/21} \\
 & Both masks & 0.069 & 147.54 & \underline{2/21} & 0.021 & \underline{188.19} & \textbf{1/21} \\
 & Pooled object & \textbf{0.245} & \textbf{137.25} & \underline{2/21} & \underline{0.123} & 201.40 & \textbf{1/21} \\
\midrule
V-JEPA~2.1-L $\times2$ & Unmasked & \underline{0.057} & 173.81 & \textbf{2/21} & \textbf{0.201} & \underline{196.85} & \textbf{0/21} \\
 & Goal mask & 0.013 & \textbf{145.88} & \underline{1/21} & 0.046 & \textbf{161.71} & \textbf{0/21} \\
 & Candidate mask & 0.014 & 215.10 & \underline{1/21} & -0.003 & 245.07 & \textbf{0/21} \\
 & Both masks & \underline{0.057} & 173.85 & \textbf{2/21} & -0.010 & 206.01 & \textbf{0/21} \\
 & Pooled object & \textbf{0.197} & \underline{164.49} & \textbf{2/21} & \underline{0.101} & 221.61 & \textbf{0/21} \\
\bottomrule
\end{tabular}
\par\smallskip
\begin{tabular}{@{}lrrrr@{}}
\toprule
& \multicolumn{2}{c}{DINOv3-L} & \multicolumn{2}{c}{V-JEPA~2-AC-g $\times2$} \\
\cmidrule(lr){2-3}\cmidrule(lr){4-5}
Objective & AF success & RV success & AF success & RV success \\
\midrule
Unmasked & \textbf{4/21} & \underline{1/21} & 1/21 & \underline{0/21} \\
Goal mask & \underline{3/21} & \textbf{2/21} & \textbf{3/21} & \underline{0/21} \\
Candidate mask & 1/21 & 0/21 & 0/21 & \underline{0/21} \\
Both masks & \underline{3/21} & \underline{1/21} & \underline{2/21} & \underline{0/21} \\
Pooled object & \underline{3/21} & \underline{1/21} & \underline{2/21} & \textbf{1/21} \\
\bottomrule
\end{tabular}
\par\smallskip
\begin{minipage}{\textwidth}\footnotesize
Success counts tasks whose selected candidate is within 15\,cm of the goal at decision 0. The benchmark supplies simulator object masks. In a camera-based system a video segmenter such as SAM~2~\cite{ravi2025sam2} could supply this input. Empty masks use full-image feature distance. $\times2$ denotes identical frames.
\end{minipage}
\end{table*}

\begin{table*}[!tp]
\centering
\caption{\textbf{Closed-loop planning with additional frozen encoders.} All rows use full-image patch distances on simulator futures. Dist.\ is mean final distance in cm and Succ.\ is tasks solved out of 21. The five-objective comparison for DINOv3-L and V-JEPA~2-AC-g is in \cref{tab:main_results}.}
\label{tab:supp_planning_frozen}
\small
\setlength{\tabcolsep}{3pt}
\renewcommand{\arraystretch}{1.04}
\begin{tabular}{@{}lrrrr@{}}
\toprule
& \multicolumn{2}{c}{Arm-free} & \multicolumn{2}{c}{Robot-visible} \\
\cmidrule(lr){2-3}\cmidrule(lr){4-5}
Encoder & Dist.\ (cm)$\downarrow$ & Succ.$\uparrow$ & Dist.\ (cm)$\downarrow$ & Succ.$\uparrow$ \\
\midrule
DINOv2-S & 49.08 & 7/21 & 47.95 & 3/21 \\
DINOv2-B & 38.11 & 7/21 & 49.69 & 3/21 \\
V-JEPA~2-L (2 frames) & 47.22 & 5/21 & 39.51 & 2/21 \\
V-JEPA~2.1-L (image) & 49.66 & 5/21 & 45.46 & 1/21 \\
V-JEPA~2.1-L (2 frames) & 50.77 & 2/21 & 44.07 & 2/21 \\
\bottomrule
\end{tabular}
\end{table*}

\setcounter{table}{20}
\begin{table*}[t]
\centering\small
\caption{\textbf{Instance-held-out final distance and regret} across ten tasks. Distances are means with sample standard deviations over three planning repeats. Regret uses the full decision-0 candidate bank. \swatch{heldbest}\,/\,\swatch{heldsecond} mark the best and second-best values in each column.}
\label{tab:supp_heldout_instances}
\setlength{\tabcolsep}{7pt}
\begin{tabular}{@{}lrrr>{\columncolor{white}[\tabcolsep][0pt]}r@{}}
\toprule
& \multicolumn{2}{c}{Final distance (cm)$\downarrow$} & \multicolumn{2}{c}{Regret (mm)$\downarrow$} \\
\cmidrule(lr){2-3}\cmidrule(lr){4-5}
Cost & Arm-free & Robot-visible & Arm-free & Robot-visible \\
\midrule
\multicolumn{5}{l}{\textit{Frozen feature distance}} \\
DINOv3-L, all patches & 37.32{\scriptsize$\,\pm\,$0.10} & 37.37{\scriptsize$\,\pm\,$0.06} & 177.87 & 132.82 \\
DINOv3-L, pooled object & 24.37{\scriptsize$\,\pm\,$0.01} & 35.87{\scriptsize$\,\pm\,$0.06} & 91.26 & 158.53 \\
V-JEPA~2-L, all patches & 34.43{\scriptsize$\,\pm\,$0.27} & 37.36{\scriptsize$\,\pm\,$0.14} & 109.78 & 153.93 \\
V-JEPA~2-L, pooled object & 34.02{\scriptsize$\,\pm\,$0.00} & 40.11{\scriptsize$\,\pm\,$0.01} & 117.76 & 165.05 \\
\midrule
\multicolumn{5}{l}{\textit{DOCK}} \\
DINOv3-L, co-trained readout & 21.09{\scriptsize$\,\pm\,$1.20} & 31.41{\scriptsize$\,\pm\,$0.15} & 54.78 & 95.55 \\
DINOv3-L, fresh readout & \hsecond 20.84{\scriptsize$\,\pm\,$1.07} & 30.73{\scriptsize$\,\pm\,$0.28} & \hsecond 49.60 & \hsecond 95.14 \\
V-JEPA~2-L, co-trained readout & 24.38{\scriptsize$\,\pm\,$0.11} & \hsecond 29.07{\scriptsize$\,\pm\,$0.28} & 52.82 & 104.05 \\
V-JEPA~2-L, fresh readout & \hbest 20.03{\scriptsize$\,\pm\,$0.27} & \hbest 28.80{\scriptsize$\,\pm\,$0.04} & \hbest 48.14 & \hbest 94.21 \\
\bottomrule
\end{tabular}
\end{table*}

\begin{table*}[t]
\centering\small
\caption{\textbf{Category-held-out final distance and regret} across nine tasks. Distances are means with sample standard deviations over three planning repeats. Regret uses the full decision-0 candidate bank. \swatch{heldbest}\,/\,\swatch{heldsecond} mark the best and second-best values in each column.}
\label{tab:supp_heldout_categories}
\setlength{\tabcolsep}{7pt}
\begin{tabular}{@{}lrrr>{\columncolor{white}[\tabcolsep][0pt]}r@{}}
\toprule
& \multicolumn{2}{c}{Final distance (cm)$\downarrow$} & \multicolumn{2}{c}{Regret (mm)$\downarrow$} \\
\cmidrule(lr){2-3}\cmidrule(lr){4-5}
Cost & Arm-free & Robot-visible & Arm-free & Robot-visible \\
\midrule
\multicolumn{5}{l}{\textit{Frozen feature distance}} \\
DINOv3-L, all patches & 43.78{\scriptsize$\,\pm\,$0.00} & 49.60{\scriptsize$\,\pm\,$0.00} & 187.82 & 160.61 \\
DINOv3-L, pooled object & 33.13{\scriptsize$\,\pm\,$0.00} & 39.85{\scriptsize$\,\pm\,$0.00} & 136.38 & 128.46 \\
V-JEPA~2-L, all patches & 54.86{\scriptsize$\,\pm\,$0.00} & 48.95{\scriptsize$\,\pm\,$0.00} & 188.50 & 143.98 \\
V-JEPA~2-L, pooled object & 41.07{\scriptsize$\,\pm\,$0.00} & 42.58{\scriptsize$\,\pm\,$0.02} & 172.06 & 119.08 \\
\midrule
\multicolumn{5}{l}{\textit{DOCK}} \\
DINOv3-L, co-trained readout & \hsecond 20.30{\scriptsize$\,\pm\,$0.00} & \hsecond 30.78{\scriptsize$\,\pm\,$0.15} & \hbest 37.56 & \hbest 75.94 \\
DINOv3-L, fresh readout & \hbest 19.58{\scriptsize$\,\pm\,$0.00} & \hbest 28.80{\scriptsize$\,\pm\,$0.00} & \hsecond 38.56 & \hsecond 89.98 \\
V-JEPA~2-L, co-trained readout & 24.61{\scriptsize$\,\pm\,$0.00} & 32.60{\scriptsize$\,\pm\,$0.08} & 77.78 & 120.08 \\
V-JEPA~2-L, fresh readout & 24.51{\scriptsize$\,\pm\,$0.00} & 31.68{\scriptsize$\,\pm\,$0.00} & 67.97 & 97.70 \\
\bottomrule
\end{tabular}
\end{table*}

\end{document}